\documentclass[sigconf]{acmart}
\usepackage{amsmath,amsfonts}
\usepackage{multirow}
\usepackage{makecell}
\AtBeginDocument{%
  }

\begin{document}

%%
%% The "title" command has an optional parameter,
%% allowing the author to define a "short title" to be used in page headers.
\title{HybridHash: Hybrid Convolutional and Self-Attention Deep Hashing for Image Retrieval}

%%
%% The "author" command and its associated commands are used to define
%% the authors and their affiliations.
%% Of note is the shared affiliation of the first two authors, and the
%% "authornote" and "authornotemark" commands
%% used to denote shared contribution to the research.

\author{Chao He}
\email{3089840275@qq.com}
\affiliation{%
	\institution{School of Computer Science, Inner Mongolia University, Provincial Key Laboratory of Mongolian Information Processing Technology, National and Local Joint Engineering Research Center of Mongolian Information Processing}
	\streetaddress{No.235 West College Road,Saihan Distric}
	\city{Hohhot}
	\country{China}
}

\author{Hongxi Wei}
\authornotemark[1]
\email{cswhx@imu.edu.cn}
\affiliation{%
	\institution{School of Computer Science, Inner Mongolia University, Provincial Key Laboratory of Mongolian Information Processing Technology, National and Local Joint Engineering Research Center of Mongolian Information Processing}
	\streetaddress{No.235 West College Road,Saihan Distric}
	\city{Hohhot}
	\country{China}
}

%%
%% By default, the full list of authors will be used in the page
%% headers. Often, this list is too long, and will overlap
%% other information printed in the page headers. This command allows
%% the author to define a more concise list
%% of authors' names for this purpose.
% \renewcommand{\shortauthors}{Trovato et al.}

%%
%% The abstract is a short summary of the work to be presented in the
%% article.
\begin{abstract}
Deep image hashing aims to map input images into simple binary hash codes via deep neural networks and thus enable effective large-scale image retrieval. Recently, hybrid networks that combine convolution and Transformer have achieved superior performance on various computer tasks and have attracted extensive attention from researchers. Nevertheless, the potential benefits of such hybrid networks in image retrieval still need to be verified. To this end, we propose a hybrid convolutional and self-attention deep hashing method known as HybridHash. Specifically, we propose a backbone network with stage-wise architecture in which the block aggregation function is introduced to achieve the effect of local self-attention and reduce the computational complexity. The interaction module has been elaborately designed to promote the communication of information between image blocks and to enhance the visual representations. We have conducted comprehensive experiments on three widely used datasets: CIFAR-10, NUS-WIDE and IMAGENET. The experimental results demonstrate that the method proposed in this paper has superior performance with respect to state-of-the-art deep hashing methods. Source code is available https://github.com/shuaichaochao/HybridHash.
\end{abstract}

%%
%% The code below is generated by the tool at http://dl.acm.org/ccs.cfm.
%% Please copy and paste the code instead of the example below.
%%
\begin{CCSXML}
	<ccs2012>
	<concept>
	<concept_id>10010147.10010178.10010224.10010225.10010231</concept_id>
	<concept_desc>Computing methodologies~Visual content-based indexing and retrieval</concept_desc>
	<concept_significance>500</concept_significance>
	</concept>
	</ccs2012>
\end{CCSXML}

\ccsdesc[500]{Computing methodologies~Visual content-based indexing and retrieval}
%%
%% Keywords. The author(s) should pick words that accurately describe
%% the work being presented. Separate the keywords with commas.
\keywords{deep hashing, image retrieval, vision Transformer, hash code}
%% A "teaser" image appears between the author and affiliation
%% information and the body of the document, and typically spans the
%% page.

%%
%% This command processes the author and affiliation and title
%% information and builds the first part of the formatted document.
\maketitle

\section{Introduction}
In recent years, with the rapid development of the Internet and the popularization of mobile media devices, the image and video data on the Web have exploded. Of these, with its intuitive and understandable characteristics, images have become a medium for carrying information, which is one of the main sources for people to obtain information. Nevertheless, the amount of images is huge, and it is a challenging task to accurately and efficiently retrieve the images people desire. Therefore, large-scale image retrieval has become one of the research hotspots  and has attracted extensive attention from researchers \cite{cui2019scalable,brogan2021fast,xia2014supervised,zhu2016deep,cao2017hashnet}. The large-scale image retrieval task supports the retrieval of relevant images from large image databases, and is broadly applied to scenarios such as search engines, recommender systems, new media software, etc. Among the numerous methods that can accomplish the large-scale image retrieval task, hashing is one of the most effective methods for image retrieval with its extremely fast speed and low memory usage \cite{liu2012supervised,zhang2010self}. It aims to learn a hash function that maps the image in the high-dimensional pixel space to the low-dimensional Hamming space, while the similarity of the images in the original space can be preserved \cite{guo2017learning}.

Existing hashing methods generally involve two phases. The first phase aims to extract image features which are mainly divided into two categories: hand-crafted based methods and deep learning based methods. Hand-crafted based methods \cite{charikar2002similarity,weiss2008spectral} learn hash functions via hand-crafted visual descriptors \cite{oliva2001modeling} (i.e., image features). However, hand-crafted features can not guarantee the semantic similarity of the raw image pairs, resulting in degraded performance in the subsequent hash function learning process. In comparison to hand-crafted based methods, deep learning based methods \cite{lin2015deep,zhu2016deep,zheng2020deep,zhang2019improved} can extract more accurate features and achieve significant performance improvements. The second phase utilizes various nonlinear functions to squeeze the image features into binary codes, and designs diverse loss functions \cite{li2015feature,cao2017hashnet,fan2020deep,yuan2020central} to guarantee the semantic similarity of the raw image pairs. 

Recently, Transformer \cite{vaswani2017attention} has achieved great success in natural language processing (NLP) \cite{devlin2018bert}. Since Transformer has robust overall modeling ability and excellent computational efficiency, some researchers tried to apply Transformer into the field of computer vision. Vision Transformer (ViT) \cite{dosovitskiy2020image} was the first model to apply Transformer into image classification tasks and achieve state-of-the-art performance. Initially the input images $ \left(224 \times 224 \times 3\right) $ are divided into 196 non-overlapping patches ( each patch has a fixed size of $ 16 \times 16 \times 3 $), which is analogous to word tokens in NLP. Then, these patches are fed into stacked standard Transformer blocks to model global relationships and extract features for classification. Inspired by the design paradigm of ViT, many variants of vision Transformers tailored for computer vision tasks have emerged. These Transformers meet and even exceed state-of-the-art convolutional neural network (CNN) based methods on a variety of computer tasks (e.g., object re-recognition \cite{he2021transreid}, semantic segmentation \cite{wang2021max}, etc.), which have motivated the exploration of the potential benefits of Transformer in image retrieval. Nevertheless, the original Vision Transformer (ViT) excels at capturing long-range dependencies, which tends to ignore local features, and weakly interacts the local feature information with the global feature information. Therefore, the existing Transformer-based image retrieval methods adopt dual-stream feature learning \cite{chen2022transhash}, multi-scale feature fusion \cite{li2023msvit} approach as to capture local features and enhance the interaction ability of local feature information with global feature information. 

In comparison to Transformer, CNNs have a more robust ability to extract local features. It has lately been demonstrated that combining multi-head self-attention in Transformers and convolutional layers in CNNs are beneficial \cite{park2022vision}. Consequently, many hybrid networks have been proposed \cite{guo2022cmt,fan2023rmt,wu2021cvt,tu2022maxvit,lee2022mpvit} by taking advantage of Transformers to capture long-range dependencies, and of CNNs extract local information, which can outperform not only canonical Transformers, but also high-performance convolutional models. Although hybrid networks combining Transformers and CNNs have achieved superior performance on various computer tasks, the potential benefits of hybrid networks for image retrieval tasks still need to be verified.

In this paper, we propose a novel hybrid network based deep hashing method called \textbf{HybridHash}. Specifically, with respect to pairwise hashing learning, we design a hybrid backbone network by utilizing Transformer to capture long-range dependencies and CNNs to extract local information, which is essentially two identical networks sharing the same parameters. Since global self-attention between pixel pairs in high-resolution images is computationally expensive, we maintain the original attention and adopt the design of the aggregation function \cite{zhang2022nested} to achieve local self-attention \cite{vaswani2021scaling}. Furthermore, we adopt the stage-wise architecture similar to CNNs \cite{guo2022cmt,he2016deep,tan2019efficientnet}, and elaborately design the interaction module to gradually decrease the resolution and flexibly increase the channel dimension. The interaction module is designed as two parallel branches. Concretely, the first branch utilizes convolutional operations to achieve local interactions between image blocks, and the second branch employs self-attention to communicate global information between image blocks. Note that image block is generated by image patches aggregation. Finally, average pooling is utilized to replace class tokens in ViT for obtaining image features, followed by hash layer to output binary codes. The major advantages of our proposed model are summarized as follows. First of all, compared to TransHash \cite{chen2022transhash}, the features generated in the first stage of HybridHash can maintain the higher resolution, i.e., $ H/4 \times W/4 $ opposed to $ H/32 \times W/32 $ in TransHash, which can preserve more detailed information. Secondly, the introduction of aggregation function can improve accuracy and data efficiency while bringing interpretability benefits. In the end, an interaction module is utilized to promote the communication of information across image patches and to enhance the visual representation. To preserve the semantic similarity of image pairs in feature space, we adopt maximum likelihood estimation to pull close similar pairs and push away dissimilar pairs in Hamming space. Since the data distribution is unbalanced in reality, weights are attached to the maximum likelihood estimation, termed as Weighted Maximum Likelihood (WML) \cite{cao2017hashnet} estimation. 

In summary, the main contributions of this paper are listed as follows:

\begin{itemize}
	\item A novel deep hashing hybrid network (HybridHash) is proposed by taking advantage of Transformers to capture long-range dependencies and of CNNs to extract local information. 
	\item The aggregation function inside HybridHash achieves the effect of local self-attention, it is thus essential to communicate information across the image blocks. We elaborately design an interaction module to promote information communication across blocks and enhance visual representation, where the convolution operation achieves local information communication across image blocks, and self-attention accomplishes overall modeling for all image blocks.  
	\item We perform comprehensive experiments on three widely-studied datasets (CIFAR-10 \cite{krizhevsky2009learning}, NUS-WIDE \cite{chua2009nus}, and IMAGENET \cite{russakovsky2015imagenet}.). Experimental results indicate that our proposed HybridHash has superior performance compared to state-of-the-art deep supervised hashing methods.
\end{itemize}

\begin{figure*}[ht]
	\centering
	\includegraphics[width=\linewidth]{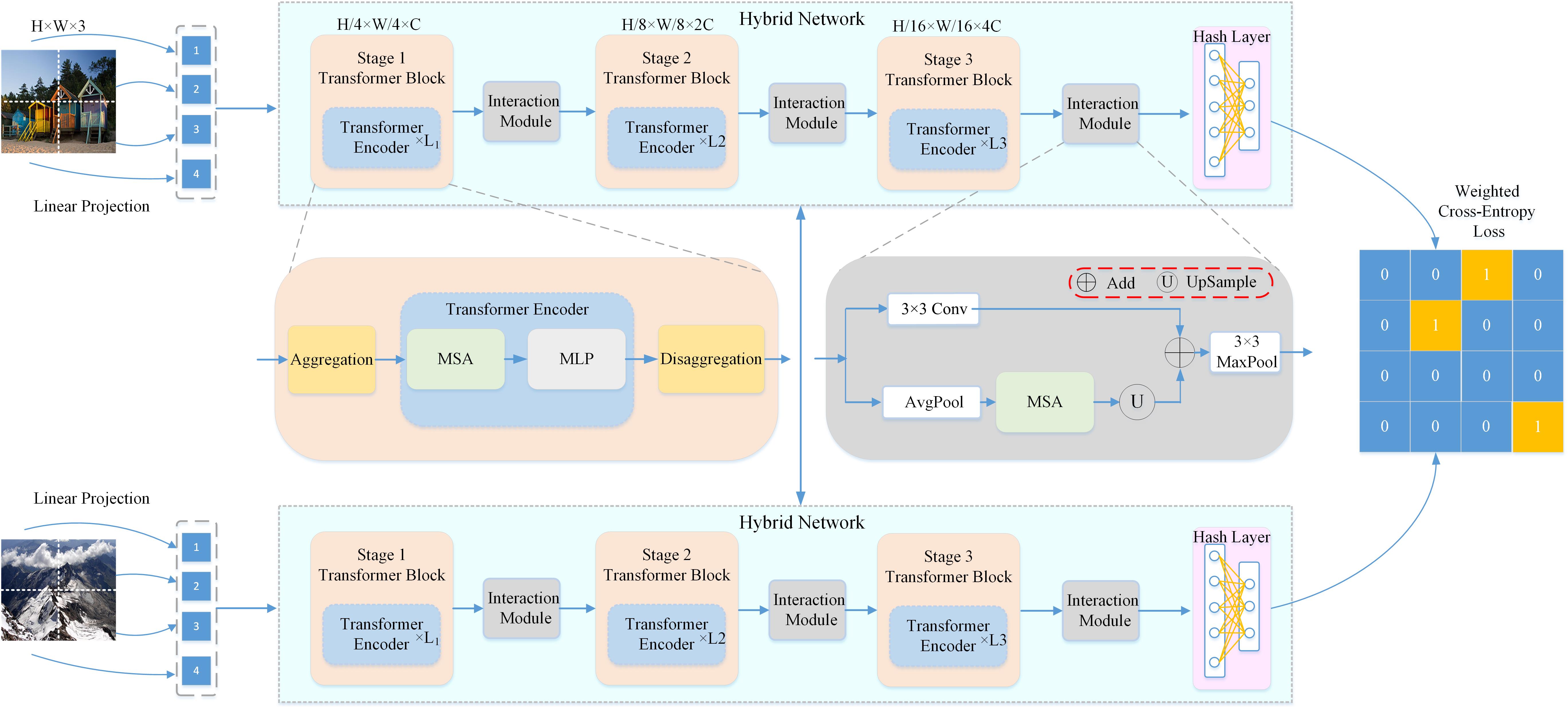}
	\caption{The detailed architecture of the proposed HybridHash. We adopt similar segmentation as ViT to divide the image with finer granularity and feed the generated image patches into the Transformer Block. The whole hybrid network consists of three stages to gradually decrease the resolution and increase the channel dimension. Interaction modules followed by each stage to promote the communication of information about the image blocks. Finally, the binary codes are output after the hash layer.}
\end{figure*}
\section{RELATED WORKS}

\subsection{General Overview of Vision Transformer}
The ViT \cite{dosovitskiy2020image} was the first model to introduce Transformer into image classification tasks and achieved superior performance. It has attracted extensive attention from researchers. The current ViT-based models have achieved excellent results in different computer vision tasks such as object detection \cite{carion2020end} and semantic segmentation \cite{wang2021max}. However, ViT has the disadvantages of large amount of parameters, high computational complexity, and weak local modeling ability. Therefore, several works have been aimed at designing vision Transformer models with simpler structure and higher computational efficiency that can simultaneously capture both global and local features. TNT \cite{han2021transformer} splits images with finer granularity to excavate features of objects at different scales and locations. Swin Transformer \cite{liu2021swin} adopts hierarchical design and utilizes the Shift-Window to capture multi-scale features by merging image patches from the bottom up. The Shift-Window scheme provides greater efficiency by limiting the self-attention computation to non-overlapping local windows while also allowing for cross-window connection. Since there are still gaps in both performance and computational cost between Transformers and existing CNNs. CMT \cite{guo2022cmt} proposes a novel Transformer-based hybrid network by taking advantage of transformers to capture long-range dependencies and of CNNs to extract local information. On the basis of the above architecture, many works \cite{zhu2023biformer,fan2023rmt,liu2023efficientvit,hatamizadeh2023fastervit,han2023flatten} have developed improvements on the self-attention of the vision Transformer to derive better computational efficiency. These state-of-the-art works based on vision Transformer bring new enlightenment to the image retrieval task.

\subsection{Deep Supervised Hashing for Image Retrieval}
To reduce storage usage and speed up retrieval, some early works \cite{liu2012supervised,lin2014fast,zhang2010self} on image retrieval introduced hashing, which map images into simple binary codes. As CNN has demonstrated superior performance in various computer vision tasks such as image classification \cite{krizhevsky2012imagenet,szegedy2015going} and object detection \cite{zhang2015improving}, many CNN-based deep supervised hashing methods have been proposed. Deep pairwise supervised hashing (DPSH) \cite{li2015feature} was the first method which utilized pairwise labels to learn feature representations and hash functions simultaneously. Deep Learning to Hash by Continuation (HashNet) \cite{cao2017hashnet} addressed the problem of ill-posed gradients when optimizing deep networks with non-smooth binary activations via a continuous method, which can exactly learn binary hash codes from imbalanced similar data. On top of HashNet, Deep Cauchy Hash for Hamming space retrieval (DCH) \cite{cao2018deep} proposed a novel pairwise cross-entropy loss based on the Cauchy distribution, that significantly penalizes similar image pairs with Hamming distances larger than a given Hamming radius threshold. Maximum margin Hamming hashing (MMHH) \cite{kang2019maximum} enabled constant time search by hash lookup in Hamming space retrieval which could promote retrieval efficiency on very large databases. Deep Polarized Network for Supervised Learning of Accurate Binary Hashing Codes (DPN) \cite{fan2020deep} further proposed a novel polarization loss. It has guaranteed to minimize the original Hamming distance-based loss without quantization error while avoiding the complex binary optimization solving. 

Motivated by recent advancements of ViT, Transhash \cite{chen2022transhash} proposes a pure Transformer-based deep hash learning framework and innovates dual-stream feature learning to learn discriminative global and local features. HashFormer \cite{li2022hashformer} further utilizes ViT as backbone network, and treats binary codes as intermediate representations of surrogate tasks (i.e., image classification) as well as proposes average precision loss. Immediately after that, MSViT \cite{li2023msvit} obtained different scale features by processing image patches with different granularity and fused them effectively. Nevertheless, the ability of ViT to learn local features is limited with respect to CNNs \cite{guo2022cmt}. Therefore, deep hashing method for hybrid CNNs and Transformers warrants investigation.

\section{HybridHash}

\subsection{Overall Architecture}
The overall architecture of HybridHash is illustrated in Figure 1, which accepts pairwise input images $ \left\{\boldsymbol{x}_i, \boldsymbol{x}_j \right\} $ and denotes the images as binary codes. HybridHash integrates three main key components called Transformer Block, Interaction module and Hash layer. 

Given an image $ \boldsymbol{x}_i \in \mathbb{R}^{H\times W\times 3} $, we first split it into a series of image patches $ \boldsymbol{X}_{ip} \in \mathbb{R}^{M\times S^2\times 3}$. $ (H, W) $ is the height and width of the image and $ (S, S ) $ is the height and width of the each image patch. $ M = HW/S^2 $ is the sequence length of the Transformer Block input. Then a linear projection layer $ \boldsymbol{E} \in \mathbb{R}^{S^2\times 3\times D} $ is used to map each image patch to an embedding in $ \mathbb{R}^D $. This way all patch embeddings are obtained, denoted as $ \boldsymbol{X}^P \in \mathbb{R}^{M \times D} $. Subsequently, $ \boldsymbol{X}^P \in \mathbb{R}^{M \times D} $ is reshaped into $ \boldsymbol{X}^P \in \mathbb{R}^{H/S \times W/S \times D} $ and is gone through three stages of the Transformer Block for hierarchical representation extraction. The Interaction module is utilized after the each stage to promote the communication of information between the image blocks and to increase the channel dimension. Finally, the predicted hash codes are output through the hash layer.

\begin{figure}[t]
	\centering
	\includegraphics[width=\linewidth]{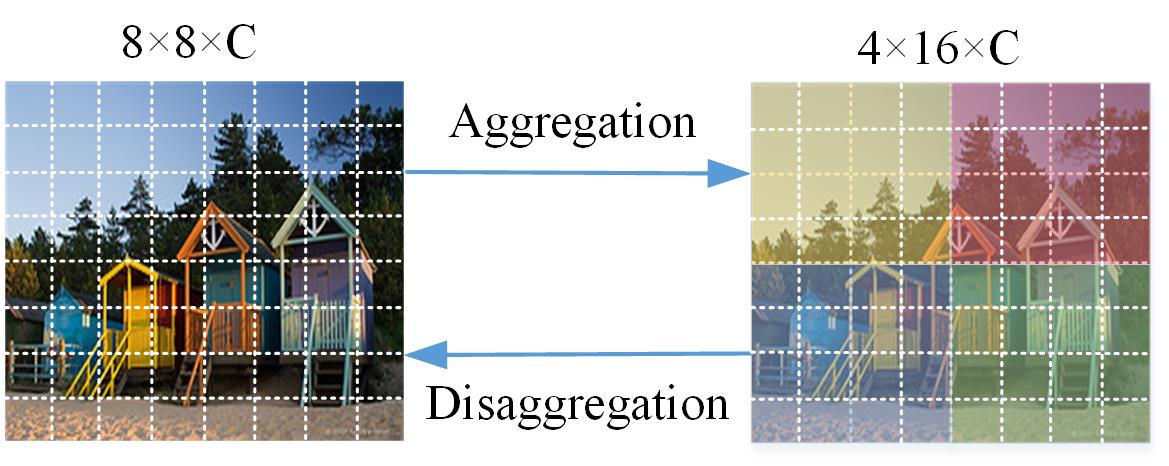}
	\caption{An instance demonstrating the detailed process of aggregation and disaggregation. For each feature map, we transform the original $ 8 \times 8 $ image patches into 4 image blocks (each block is represented by different color) via the aggregation function, and each image block contains 16 image patches. Self-attention is performed exclusively within each image block.}
\end{figure}

\subsection{Transformer Block}
Transformer Block consists of three main key components which are Aggregation, Transformer Encoder and Disaggregation.

\textit{Aggregation and Disaggregation Function}. For the purpose of generating features with higher resolution in the first stage to preserve more detailed information, we set the size of the image patch to 4, i.e. $ \boldsymbol{X}^P \in \mathbb{R}^{H/4 \times W/4 \times D} $.  Then we separate all patch embeddings $ \boldsymbol{X}^P $ into blocks to generate the inputs $ \boldsymbol{Y} \in \mathbb{R}^{T_n \times n \times D} $, where $ T_n $ is the total number of blocks and $ n $ is the sequence length inside each image block. In this way, self-attention can be performed in each image block to extract local features. The disaggregation module performs the opposite process as described above, which reshapes $ \boldsymbol{Y}$  into $ \boldsymbol{X} $ for feeding the interaction module. We demonstrate an instance to illustrate the detailed process of aggregation and disaggregation, as shown in Figure 2. 

\textit{Transformer Encoder}. The Transformer Encoder consists of multi-head self-attention (MSA) \cite{vaswani2017attention}, and a fully connected network (MLP) with skip connections \cite{he2016deep} and LayerNorm (LN) \cite{ba2016layer}. The MLP contains two layers with a GELU non-linearity. We stack multiple Transformer Encoders within each block and perform multi-head self-attention. In this way, the Transformer Encoder can process all blocks in a parallel manner, which will increase the training speed of the network. Before being fed to the Transformer Encoder, the trainable position embedding vectors $ \boldsymbol{E}_{pos} \in \mathbb{R}^{T_n \times n \times D} $ are added to $ \boldsymbol{Y} $. 

\begin{equation}
	Multiple\times \left\{
	\begin{aligned}
		Y^*={\rm MSA}\left({\rm LN}\left(Y\right)\right)+Y  \\
		Y={\rm MLP}\left({\rm LN}\left(Y^*\right)\right)+Y^*
	\end{aligned}
	\right.
\end{equation}

\subsection{Interaction Module}
Since self-attention is performed only within each image block, the communication of information between blocks takes on great importance. In this work, we design a novel interaction module that aims to efficiently achieve the communication of information between image blocks in the simplest structure. Figure 3 demonstrates the detailed structure of the interaction module. Specifically, the interaction module is composed of two branches. The first branch only comprises $ 3 \times 3 $ convolution. 

\begin{equation}
   \boldsymbol{X}_{local} = \mathrm{Conv}_{3 \times 3}\left(\boldsymbol{X}^P\right)
\end{equation}
There are two major roles for the first branch, one is to further extract image block features and mine local information, and the other is to yield communication of information at the edges of the image block. For the second branch, we introduce the notion of \textit{block tokens} (BTs), which obtain large attention footprints at low cost and play the summarizing role of the entire image blocks. To begin with, we initialize BTs by pooling to $ A=4^b $ tokens per image block.

\begin{equation}
	\begin{aligned}
		\hat{\boldsymbol{X}}_{B} & = \mathrm{Conv}_{1 \times 1}\left(\boldsymbol{X}^P\right)	\\
		\hat{\boldsymbol{X}}_{BT} & = \mathrm{AvgPool}_{HW/S^2 \rightarrow T_nA}\left(\hat{\boldsymbol{X}}_{B}\right)
	\end{aligned}
\end{equation}
where $ \hat{\boldsymbol{X}}_{BT} $ and AvgPool denote the block tokens and feature pooling operation, respectively. $ b $ is set to 0, but can be changed to control the amount of BTs. The current approach with conv+pooling gives flexibility with the image size. These pooled tokens represent the summaries of their respective image blocks, and we let $ A \ll n $. The initialization process of BT is performed only once in each interaction module. Note that each image block has a unique set of block tokens. Subsequently, the trainable position embedding vectors $ \boldsymbol{E}_{pos-bt} \in \mathbb{R}^{ T_nA \times D} $ are added to $ \hat{\boldsymbol{X}}_{BT} $, BTs undergo the attention procedure:
\begin{equation}
	\hat{\boldsymbol{X}}_{BT} = \mathrm{MSA}\left(\hat{\boldsymbol{X}}_{BT}+\boldsymbol{E}_{pos-bt}\right)
\end{equation}
where MSA represents multi-head self-attention. Next, we utilized bilinear interpolation to upsample the block tokens (BTs) for the purpose of recovering the original feature sizes. In this way, global communication of information between image blocks is achieved and global features of the image are also captured. 

\begin{equation}
	\boldsymbol{X}_{global} = \mathrm{UpSample}\left(\hat{\boldsymbol{X}}_{BT}\right)
\end{equation}
Finally, the local and global features are fused followed by LayerNorm (LN) and $ 3 \times 3 $ max pooling to reduce the feature resolution and fed into the second stage of the network. 

\begin{equation}
	\boldsymbol{X}^P = \mathrm{MaxPool}\left( \mathrm{LN}\left(\boldsymbol{X}_{global}+\boldsymbol{X}_{local}\right)\right)
\end{equation}
Where $ \boldsymbol{X}^P \in \mathbb{R}^{H/8 \times W/8 \times D^{'}}$. The total number of image blocks $ T_n $ is reduced by a factor of 4 until reduced to 1 at the top, while the sequence length $ n $ always remains the same. $ D^{'} \geq D$ depends on the specific model configuration. The Hash Layer transforms image features into hash codes, which contains one layer with a TANH non-linearity.

\begin{figure*}[ht]
	\centering
	\includegraphics[width=\linewidth]{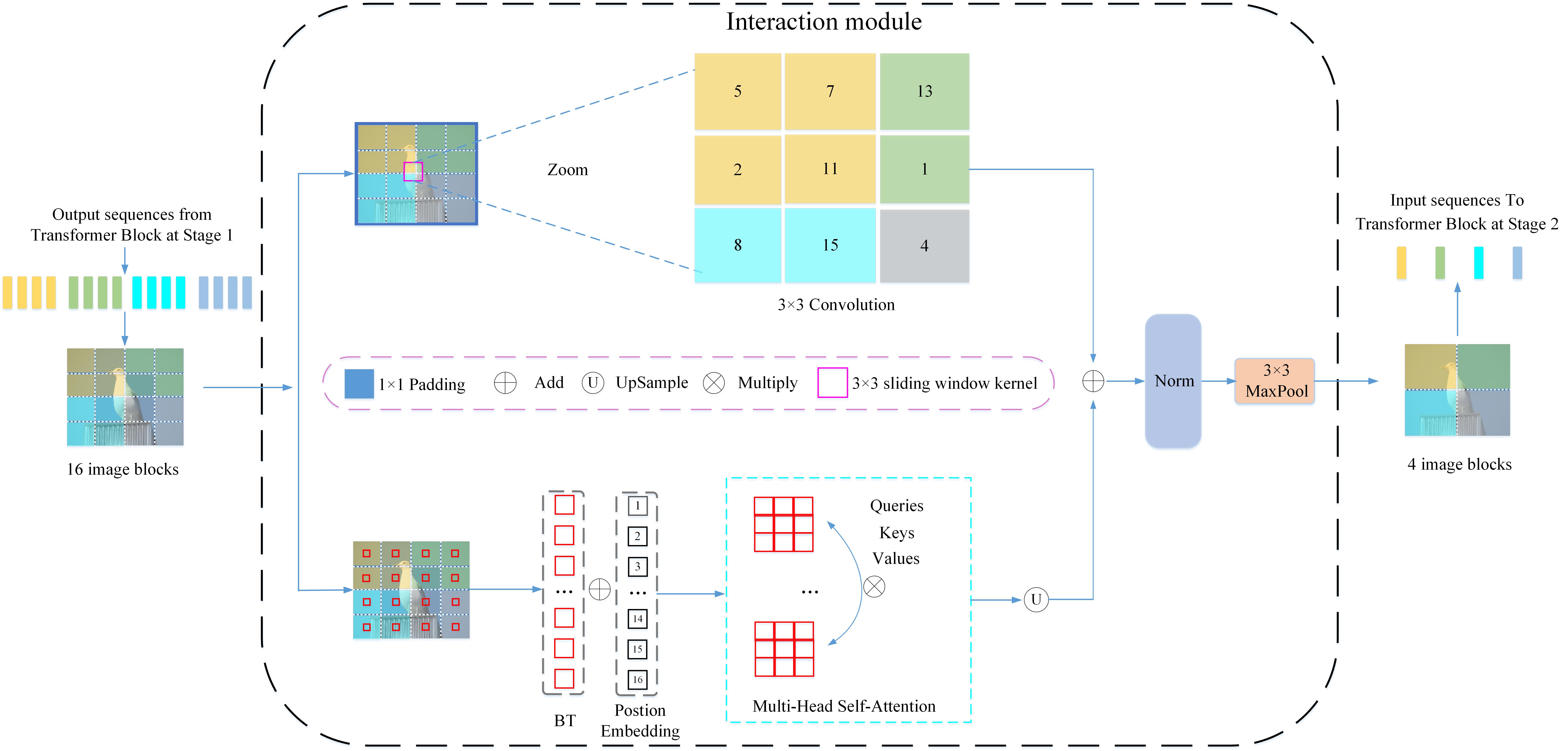}
	\caption{Illustration of Interaction Module. Specifically, the interaction module incorporates two branches, the first one utilizes convolutional operations to achieve local communication of information across image blocks, and the second one utilizes block tokens to accomplish global communication of information between image blocks. Finally, the two features are fused.}
\end{figure*}

\subsection{Weighted Cross-entropy Loss}
To preserve the similarity information of pairwise images and simultaneously learn deep hashing from unbalanced data, we adopt Weighted Maximum Likelihood to generate binary hash codes \cite{cao2017hashnet}. Given training
images ($ \boldsymbol{h}_i, \boldsymbol{h}_j,  s_{ij} $). Here, $ S = \left\{s_{ij}\right\}$ represents the similarity matrix where $ s_{ij}=1 $ if $ \boldsymbol{h}_i $ and $ \boldsymbol{h}_j $ are from the same class and $ s_{ij}=0 $ otherwise. For all $ N $ training images, the Weighted Maximum Likelihood (WML) estimation of the hash codes $ \boldsymbol{H} = \left\{\boldsymbol{h}_1, \ldots \boldsymbol{h}_N\right\} $ is

\begin{equation}
    \mathrm{log}P\left(S|\boldsymbol{H}\right) = \sum_{s_{ij} \in S}w_{ij}\mathrm{log}\left(s_{ij}|h_i,h_j\right)
\end{equation}
Where $ \mathrm{log}P\left(S/\boldsymbol{H}\right) $ is the weighted likelihood function and $ w_{ij} $ is the weight of each training pair, which is weighted in accordance with the importance of training pair misclassification to solve the problem of data imbalance \cite{dmochowski2010maximum}. Since there are only two cases of similar and dissimilar labels in $ S $, taking into account the data imbalance between similar and dissimilar pairs, we set

\begin{equation}
	w_{ij}=c_{ij}\times \left\{
	\begin{aligned}
		|S|/|S_1|, &&&&&& s_{ij}=1\\
	    |S|/|S_0|, &&&&&&  s_{ij}=0
	\end{aligned}
	\right.
\end{equation}
Where $\boldsymbol{S}_1=\left\{s_{ij}\in \boldsymbol{S}:s_{ij}=1\right\}$ is the set of similar pairs
and $\boldsymbol{S}_0=\left\{s_{ij}\in \boldsymbol{S}:s_{ij}=0\right\}$ is the set of dissimilar pairs; $c_{ij}$ is continuous similarity, i.e. $c_{ij}=\dfrac{\boldsymbol{y}_{i}\cap \boldsymbol{y}_j}{\boldsymbol{y}_{i}\cup \boldsymbol{y}_j}$ if labels $\boldsymbol{y}_i$ and $\boldsymbol{y}_j$ of $\boldsymbol{x}_i$ and $\boldsymbol{x}_j$ are given, $c_{ij}=1$ if only $s_{ij}$ is given. For an arbitrary pair $ \boldsymbol{h}_i.\boldsymbol{h}_j $, $ P\left(s_{ij}|\boldsymbol{h}_i,\boldsymbol{h}_j\right) $ is the conditional probability function of $ s_{ij}$ for a given pair of hash codes $ \boldsymbol{h}_i $ and $ \boldsymbol{h}_j $. We can define $ P\left(s_{ij}|\boldsymbol{h}_i,\boldsymbol{h}_j\right) $ as a Bernoulli distribution:

\begin{equation}
\begin{aligned}
	P\left(s_{i j} \mid \boldsymbol{h}_{i}, \boldsymbol{h}_{j}\right) & =\left\{\begin{array}{ll}
		\sigma\left(\left[\left(\boldsymbol{h}_{i}, \boldsymbol{h}_{j}\right)\right]\right), & s_{i j}=1 \\
		1-\sigma\left(\left[\boldsymbol{h}_{i}, \boldsymbol{h}_{j}\right]\right), & s_{i j}=0
	\end{array}\right. \\
	& =\sigma\left(\left[\boldsymbol{h}_{i}, \boldsymbol{h}_{j}\right]\right)^{s_{i j}}\left(1-\sigma\left(\left[\boldsymbol{h}_{i}, \boldsymbol{h}_{j}\right]\right)\right)^{1-s_{i j}}
\end{aligned}
\end{equation}
where $ \left[*\right] $ denotes the inner product and $ \sigma\left(x\right) = 1/\left(1+e^{-\alpha x}\right)$ is the adaptive sigmoid function with hyper-parameter $  \alpha $ to control its bandwidth. Since the Sigmoid function with larger $ \alpha $ will generate larger saturation zone, where the gradient in the saturation zone is zero. To perform more efficient backpropagation, we generally set $ \alpha < 1 $, which is more efficient than the typical setting of $ \alpha = 1 $. The inner product $ \left[\boldsymbol{h}_{i}, \boldsymbol{h}_{j}\right] $ and the Hamming distance $ \mathcal{H}\left(\boldsymbol{h}_{i}, \boldsymbol{h}_{j}\right) $ have a nice relationship, i.e., $ \mathcal{H}\left(\boldsymbol{h}_{i}, \boldsymbol{h}_{j}\right)=\dfrac{1}{2}\left(K-\left[\boldsymbol{h}_{i}, \boldsymbol{h}_{j}\right]\right) $. $ K $ denotes the length of the hash code. Therefore, we can observe that the smaller the Hamming distance $ \mathcal{H}\left(\boldsymbol{h}_{i}, \boldsymbol{h}_{j}\right) $ is, the larger the inner product $ \left[\boldsymbol{h}_{i}, \boldsymbol{h}_{j}\right] $is, as well as the larger the conditional probability $ P\left(1 \mid \boldsymbol{h}_{i}, \boldsymbol{h}_{j}\right) $ will be, which indicates that the $ \boldsymbol{h}_{i} $ and $ \boldsymbol{h}_{j} $ correspond to be classified as similar; otherwise, the larger the conditional probability $ P\left(0 \mid \boldsymbol{h}_{i}, \boldsymbol{h}_{j}\right) $ is, and the $ \boldsymbol{h}_{i} $  and $ \boldsymbol{h}_{j} $  correspond to be classified as dissimilar. By taking Equation (9) into the WML estimation in Equation (7), the final optimization problem can be obtained,

\begin{equation}
	\mathop{min}\limits_{\theta} \sum_{s_{ij}\in \boldsymbol{S}} w_{ij}\left(\mathrm{log}\left(1+\mathrm{exp}\left(\alpha\left[\boldsymbol{h}_i,\boldsymbol{h}_j\right]\right)\right)-\alpha s_{ij}\left[\boldsymbol{h}_i,\boldsymbol{h}_j\right]\right)
\end{equation}
Where $\theta$ denotes the set of all parameters in deep neural networks and $\alpha$ is a hyper-parameter.

\begin{table*}[t]
	\centering
	\caption{The corresponding results (MAP) on the three benchmark datasets.}
	\label{tab:commands}
	\begin{tabular}{c|cccccccccccc}
		\hline
		Datasets & \multicolumn{4}{c}{CIFAR-10@54000} & \multicolumn{4}{c}{NUS-WIDE@5000} & \multicolumn{4}{c}{IMAGENET@1000} \\
		\hline
		Methods & 16 bits & 32 bits & 48 bits & 64 bits & 16 bits & 32 bits & 48 bits & 64 bits & 16 bits & 32 bits & 48 bits & 64 bits \\
		\hline
		 SH \cite{weiss2008spectral} & - & - & - & - & 0.4058 & 0.4209 & 0.4211 & 0.4104 & 0.2066 & 0.3280 & 0.3951 & 0.4191
 \\
		 ITQ \cite{gong2012iterative}& - & - & - & - & 0.5086 & 0.5425 & 0.5580 & 0.5611 & 0.3255 & 0.4620 & 0.5170 & 0.5520\\
		 KSH \cite{liu2012supervised}& - & - & - & - & 0.3561 & 0.3327 & 0.3124 & 0.3368 & 0.1599 & 0.2976 & 0.3422 & 0.3943\\
		 BRE \cite{kulis2009learning}& - & - & - & - & 0.5027 & 0.5290 & 0.5475 & 0.5546 & 0.0628 & 0.2525 & 0.3300 & 0.3578\\
		 \hline
		 DSH \cite{liu2016deep}& 0.6145 & 0.6815 & 0.6828 & 0.6910 &0.6338 &0.6507 &0.6664 &0.6856 &0.4025 &0.4914 &0.5254 &0.5845 \\
		 DHN \cite{zhu2016deep}& 0.6544 &0.6711 &0.6921 &0.6737 &0.6471 &0.6725 &0.6981 &0.7027 &0.4139 &0.4365 &0.4680 &0.5018 \\
         DPSH \cite{li2015feature}& 0.7230 &0.7470 &0.7550 &0.7750 & 0.7156 &0.7302 &0.7426 &0.7172 & 0.4531 &0.4836 &0.5020 &0.5330\\
         HashNet \cite{cao2017hashnet}& 0.7321 &0.7632 &0.7820 &0.7912 &0.6612 &0.6932 &0.7088 &0.7231 &0.4385 &0.6012 &0.6455 &0.6714 \\
         DCH \cite{cao2018deep}&0.7562 &0.7874 &0.7929 &0.7935 &0.7012 &0.7345 &0.7306 &0.7151 &0.4356 &0.5663 &0.5872 &0.5688 \\
         MMHH \cite{kang2019maximum}& 0.7956 &0.8087 &0.8152 &0.8178 & 0.7687 &0.7874 &0.7801 &0.7514 & - & - & - & - \\
         DPN \cite{fan2020deep}&0.8250 &0.8380 &0.8300 &0.8290 & - & - & - & - & 0.6840 &0.7400 &0.7560 &0.7610\\
         TransHash \cite{chen2022transhash}& 0.9075 &0.9108 &0.9141 &0.9166 &0.7263 &0.7393 &0.7532 &0.7488 &0.7852 &0.8733 &0.8932 &0.8921 \\
         HashFormer \cite{li2022hashformer}&0.9121 &0.9167 &0.9211 &0.9236 &0.7317 &0.7418 &0.7592 &0.7597 &0.7791 &\textbf{0.8962} &0.9007 &0.9010 \\
         MSViT-B \cite{li2023msvit}&0.8982 &0.9281 &0.9380 &0.9443 & - & - & - & - &0.7869 &0.8635 &0.8926 &0.9036 \\
         HybridHash(ours) &\textbf{0.9367}  &\textbf{0.9413}   &\textbf{0.9468}  &\textbf{0.9513} &\textbf{0.7785}   &\textbf{0.7986}    &\textbf{0.8068}   &\textbf{0.8164} &\textbf{0.8028}  &0.8886   &\textbf{0.9094}  &\textbf{0.9110} \\
         
		\hline
	\end{tabular}
\end{table*}

\section{EXPERIMENTS}

\subsection{Datasets and Evaluation Protocols}
We conducted experiments on three widely used datasets for image retrieval, including CIFAR-10, NUS-WIDE and IMAGENET.

\textbf{CIFAR-10}: CIFAR-10 is a single-labeled dataset containing 60,000 images, of which 50,000 images are utilized for training and 10,000 images for testing. The dataset is available in 10 categories and each category contains 6000 images. We follow the same setup as the experiments in \cite{chen2022transhash}, where 1000 images are treated as the query set, 5000 images randomly selected from the dataset are served as the training set, and the other 54000 images are used as the retrieval (database) set. 

\textbf{NUS-WIDE}: NUS-WIDE is a multi-labeled dataset typically utilized for large-scale image retrieval tasks. The dataset is organized in 81 categories which contains 269648 images. We followed the experimental setup of \cite{chen2022transhash} and randomly selected 5000 images for testing and the other images as the retrieval (database) set. Then 10,000 images are randomly selected from the retrieval (database) set for training. 

\textbf{IMAGENET}: IMAGENET is the benchmark image dataset for the Large Scale Visual Recognition Challenge (ILSVRC 2015). Concretely, we follow the experimental setup in \cite{chen2022transhash} and randomly select 100 categories of images. All training images of these 100 categories are treated as retrieval sets, and the test images are regarded as query sets. Eventually, 100 images from each category are randomly selected as the training set. 

We adopt the mean average precision (MAP) of different bits $ \left\{16, 32, 48, 64\right\} $ to evaluate the quality of the retrieved images. Specifically, we followed similar work \cite{chen2022transhash,cao2017hashnet} and the MAP results were calculated based on the top 54,000 returned samples from the CIFAR-10 dataset, 5,000 returned samples from the NUS-WIDE dataset, and 1,000 returned samples from the IMAGENET dataset.

\subsection{Implementation Details}
All images are initially resized to $ 256 \times 256 $. For the training images, we adopt standard image augmentation techniques comprising random horizontal flipping and random cropping with cropping size 224. To speed up the convergence of the model, we utilize RMSProp as the optimizer. For the experimental parameters, the batch size is $ 64 $, the learning rate is tuned in the range of $ \left[2.5 \times 10^{-5}, 5 \times 10^{-5}\right] $ and the weight decay parameter is set to $ 10^{-5} $. We obtain the hyper-parameter $ \alpha $ of HybridHash by cross-validation as 0.5. All experiments are conducted with one Tesla V100S GPU.

\subsection{Experimental Results and Analysis}
In this section, we compare the performance of our proposed HybridHash with state-of-the-art deep hashing methods. Specifically, the competing methods can be summarized into two categories: hand-crafted based hashing methods and deep learning based hashing methods. For hand-crafted based hashing methods, we select the more frequently compared methods SH \cite{weiss2008spectral}, ITQ \cite{gong2012iterative}, KSH \cite{liu2012supervised}, and BRE \cite{kulis2009learning} for detailed comparison. For deep learning based hashing methods, we further include DSH \cite{liu2016deep}, which is one of the first works for deep convolutional neural networks dealing with the image retrieval hashing problem. Moreover, we incorporate other state-of-the-art deep hashing methods, including DHN \cite{zhu2016deep}, DPSH \cite{li2015feature}, HashNet \cite{cao2017hashnet}, DCH \cite{cao2018deep}, MMHH \cite{kang2019maximum}, DPN \cite{fan2020deep}, TransHash \cite{chen2022transhash}, HashFormer \cite{li2022hashformer}, and MSViT \cite{li2023msvit}. 

It should be emphasized that all non-deep learning methods and DSH, DHN, DPN, TransHash, where the results are derived from \cite{chen2022transhash}. Whereas HashFormer and MSViT are derived from \cite{li2022hashformer} and \cite{li2023msvit}. The results of the other methods we obtained by conducting experiments based on the open-source code provided in the original paper and maintaining the same hyper-parameters and preprocessing techniques. 

Table 1 demonstrates the MAP results of the different hashing methods on three benchmark datasets. It is obvious seen that our proposed HybridHash has remarkable performance advantage when compared to the non-deep hashing methods. The reason for the undesirable performance of the non-deep hashing methods may be that the hand-crafted visual descriptors are inaccurate, resulting in the generation of sub-optimal hash codes. While the deep hashing methods exhibit superior performance on three benchmark datasets for different hash bit lengths. Notwithstanding, our method still outperforms all competing methods. The reasons for this are twofold. First, HybridHash adopts stage-wise architecture, which can excavate the features of objects at different scales and locations. The second one is that the interaction module promotes the communication of information between image blocks and enhances the visual representations. Our method also consistently outperforms competing methods on NUSWIDE for different hash bit lengths with significant performance improvement. This indicates that HybridHash is also suitable for multi-label image retrieval, where each image contains multiple labels.

\begin{table}[t]
	\caption{MAP results for HybridHash and its variables on three
		benchmark datasets.}
	\label{tab:freq}
	\begin{tabular}{c|c|cccc}
		\hline
		Dataset & Bits & Base & Base-C & Base-M & HybridHash\\ 
		\hline
		\multirow{4}{*} {\makecell[c]{CIFAR-10\\@54000}}
		& 16  & 0.7152 & 0.9209  & 0.9262 &  \textbf{0.9367} \\
		& 32  & 0.7213 &  0.9283  & 0.9274 &  \textbf{0.9413} \\
		& 48  & 0.7465 & 0.9391   & 0.9376 & \textbf{0.9468} \\
		& 64  & 0.7534 & 0.9403   & 0.9434 & \textbf{0.9513} \\
		
		\hline
		\multirow{4}{*}{\makecell[c]{NUS-WIDE\\@5000}} 
		& 16  & 0.6167 &  0.7657  &0.7670 &  \textbf{0.7785} \\
		& 32  & 0.6284 &  0.7882  & 0.7901&  \textbf{0.7986} \\
		& 48  & 0.6041 & 0.8024   & 0.7974& \textbf{0.8068} \\
		& 64  & 0.6599 & 0.8062   & 0.8040& \textbf{0.8164} \\
		
		\hline
		\multirow{4}{*}{\makecell[c]{IMAGENET\\@1000}} 
		& 16  & 0.5345 &  0.7426  &0.7513 &  \textbf{0.8028} \\
		& 32  & 0.5753 &  0.8765   &0.8587 &  \textbf{0.8886} \\
		& 48  & 0.5729 & 0.8952   & 0.8849& \textbf{0.9094} \\
		& 64  & 0.5928 & 0.9054   & 0.8963& \textbf{0.9110} \\
		
		\hline
	\end{tabular}
\end{table}

\begin{table}[b]
	\caption{Analysis of the effects of A on three benchmark datasets.}
	\label{tab:freq}
	\begin{tabular}{c|c|cc}
		\hline
		Dataset & Bits  & A=1 & A=4 \\ 
		\hline
		\multirow{4}{*}{CIFAR-10@54000} 
		& 16      &  \textbf{0.9367} & 0.9300\\
		& 32     &  \textbf{0.9413} &  0.9352 \\
		& 48     & \textbf{0.9468} & 0.9436\\
		& 64    & \textbf{0.9513} & 0.9512 \\
		
		\hline
		\multirow{4}{*}{NUS-WIDE@5000} 
		& 16     &  \textbf{0.7785}  &  0.7753\\
		& 32    &  \textbf{0.7986} &  0.7939\\
		& 48     & \textbf{0.8068} & 0.8034\\
		& 64     & \textbf{0.8164} & 0.8125\\
		
		\hline
		\multirow{4}{*}{IMAGENET@1000} 
		& 16      &  \textbf{0.8028} &  0.7950\\
		& 32     &  \textbf{0.8886} &  0.8853\\
		& 48     & \textbf{0.9094} & 0.9089\\
		& 64     & \textbf{0.9110} & 0.9105\\
		
		\hline
	\end{tabular}
\end{table}

\subsection{Model Settings}
HybridHash accepts input images of size $ 224 \times 224 $ and slices the images with patch size $ 4 \times 4 $. To be specific, the overall model is composed of three stages with the number of stacked standard Transformer encoders as ($ L_1 $) 2, ($ L_2 $) 2, and ($ L_3 $) 15. The three stages respectively own 16, 4, and 1 image blocks, and the sequence length within each block is $ 14 \times 14 $. We set the number of self-attention heads for the three stages to 4, 8, and 16, while the hidden feature dimension is 128, 256, and 512. It is worth noting that the number of self-attention heads in the interaction module is 8.

\begin{table}[t]
	\caption{Comparison of MAP results and efficiency of different backbone networks on three benchmark datasets.}
	\label{tab:freq}
	\begin{tabular}{c|c|cccc}
		\hline
		Dataset & Bits & ResNet101 & ViT-B/32 & ViT-L/32 & Hybrid\\ 
		\hline
		\multirow{4}{*} {\makecell[c]{CIFAR-10\\@54000}}
		& 16  & 0.8631 & 0.8125  & 0.8311 &  \textbf{0.9367} \\
		& 32  & 0.8821 & 0.8278   & 0.8503 &  \textbf{0.9413} \\
		& 48  & 0.8903 & 0.8462   & 0.8836 & \textbf{0.9468} \\
		& 64  & 0.8924 & 0.8741   & 0.8932 & \textbf{0.9513} \\
		
		\hline
		\multirow{4}{*}{\makecell[c]{NUS-WIDE\\@5000}} 
		& 16  & 0.7542 & 0.6374   &0.6667 &  \textbf{0.7785} \\
		& 32  & 0.7731 & 0.6541   & 0.6835&  \textbf{0.7986} \\
		& 48  & 0.7802 & 0.6587   & 0.6791& \textbf{0.8068} \\
		& 64  & 0.7929 & 0.6832  & 0.7065& \textbf{0.8164} \\
		
		\hline
		\multirow{4}{*}{\makecell[c]{IMAGENET\\@1000}} 
		& 16  & 0.7681 & 0.6742  &0.7034 &  \textbf{0.8028} \\
		& 32  & 0.7765 & 0.7454  &0.7753 &  \textbf{0.8886} \\
		& 48  & 0.8153 & 0.7689   & 0.8021& \textbf{0.9094} \\
		& 64  & 0.8395 & 0.7802  & 0.8198& \textbf{0.9110} \\
		
		\hline
	\end{tabular}
\end{table}

\begin{table}[t]
	\caption{Comparison of the efficiency of different backbone networks on three benchmark datasets.}
	\label{tab:freq}
	\begin{tabular}{c|cc}
		\hline
		backbone & \#param & FLOPs \\ 
		\hline
		ResNet101 & 42.53M & 15.71G \\
		ViT-B/32 & 87.47M & 4.41G \\
		ViT-L/32 & 292.93M & 14.76	G \\
		Hybrid(Ours) & 55.57M & 14.17G \\
		\hline
	\end{tabular}
\end{table}

\subsection{Ablation Study}
To further analyze the overall design of our proposed method, a detailed ablation study is performed to illustrate the effectiveness of each component. Specifically, we investigated two variants of HybridHash:
\begin{itemize}
	\item \textbf{Base}: A variant without the interaction module that only utilizes $3 \times $3 max pooling for down-sampling. 
	\item \textbf{Base-C}: A variant that merely adopts the convolutional layer of $3 \times $3 with stride length 2 in the interaction module.
	\item \textbf{Base-M}: A variant that only adopts a $ 3 \times 3 $ convolutional layer and a $ 3 \times 3 $ max pooling layer in the interaction module. 
\end{itemize}

Table 2 exhibits the MAP results of HybridHash and its variants on three benchmark datasets. As can be viewed in Table 2, the removal of the interaction module leads to significant performance decreases, which demonstrates the critical importance of communicating information between image blocks. Meanwhile, it can be observed that \textbf{Base-C} and \textbf{Base-M} have approaching performance when local information is communicated between image blocks by only utilizing convolution. Nevertheless, when the global information communication branch of the interaction module is removed, we experience noticeable performance decrease on three datasets. It indicates that the interaction module we designed adequately promotes the communication of information between image blocks and enhances the visual representations. 

We further performed an ablation study on the sensitivity of the number of block tokens (BTs) $ A  $ within each image block. Since we set the size of each image block to $ 14 \times 14$, 1 or 4 block tokens (BTs) can be obtained after $ 14 \times 14 $ global average pooling or $ 7 \times 7 $ global average pooling, i.e., $  A = 1 $ or $ A = 4 $. Table 3 demonstrates the effect of $ A  $ on three benchmark datasets. It can be observed that the performance decreases when the number of block tokens increases. It is probably caused by the fact that multiple block tokens generate redundant features. According to the above observation, we empirically set $ b $ to 0 ($ A =1$) at four different hash bit lengths. 

We also compared performance and efficiency with mainstream backbone networks. Comprehensive Tables 4 and 5 observe that our proposed method utilizes less computational and parameters to obtain superior performance.

\section{CONCLUSION}
To effectively accomplish large-scale image retrieval tasks, this paper proposes a deep hashing method with hybrid convolution and self-attention (HybridHash). Specifically, HybridHash adopts stage-wise architectural design to reduce computational complexity and simultaneously learn more fine-grained features. On this basis, we elaborately design the interaction module to enable image blocks for local communication of information by utilizing convolution and to model the overall of the entire image blocks by utilizing self-attention. A weighted maximum likelihood estimation is employed for similarity preserving learning on top of all pairwise features. The entire framework is optimized in an end-to-end fashion. We have conducted extensive experiments on three benchmark datasets and the experimental results demonstrate that the method proposed in this paper indicates  superior results compared to existing state-of-the-art deep hashing methods.

%%
%% The next two lines define the bibliography style to be used, and
%% the bibliography file.
\bibliographystyle{ACM-Reference-Format}
\bibliography{sample-base}

\end{document}